\documentclass[runningheads]{llncs}
\usepackage{graphicx}
\usepackage{longtable}
\usepackage{multirow}
\usepackage{diagbox}
\usepackage{dirtytalk}
\usepackage{geometry}
\usepackage{graphicx}
\usepackage{setspace}
\usepackage{amsfonts}
\usepackage{longtable}
\usepackage{slashbox}
\usepackage{subfigure}
\usepackage{epsfig}
\usepackage{url}
\usepackage{comment}
\begin{document}
%
\title{Feature Engineering in Learning-to-Rank for Community Question Answering Task}
%
%
\author{Nafis Sajid
\and
Md. Rashidul Hasan
\and
Muhammad Ibrahim*
}
\authorrunning{N. Sajid et al.}
%
\institute{Department of Computer Science and Engineering\\ University of Dhaka, Dhaka-1000, Bangladesh\\
\email{nafissajidns@gmail.com,
mamun02inf@gmail.com,
ibrahim313@du.ac.bd*}
}
\maketitle              
\begin{abstract}
Community question answering (CQA) forums are Internet-based platforms where users ask questions about a topic and other expert users try to provide solutions. Many CQA forums such as Quora, Stackoverflow, Yahoo!Answer, StackExchange exist with a lot of user-generated data. These data are leveraged in automated CQA ranking systems where similar questions (and answers) are presented in response to the user's query. In this work, we empirically investigate a few aspects of this domain. Firstly, in addition to traditional features like TF-IDF, BM25 etc., we introduce a BERT-based feature that captures the semantic similarity between the question and answer. Secondly, most of the existing research works have focus on features extracted only from the question part; features extracted from answers have not been explored extensively. We combine both types of features in a linear fashion. Thirdly, using our proposed concepts, we conduct an empirical investigation with different rank-learning algorithms, some of which have not been used so far in CQA domain. On three standard CQA datasets, our proposed framework achieves state-of-the-art performance. We also analyze importance of the features we use in our investigation. This work is expected to guide the practitioners to select a better set of features for the CQA retrieval task.

\end{abstract}
\section{Introduction}
\label{sec:introduction motivation}

The task of Information Retrieval (IR) is to fetch relevant documents from a pool of documents with respect to some query. There may be thousands of documents containing the exact query words, but all of them are not equally relevant to the search query. Showing the most relevant document on top of the ranked document list is called document ranking. When a user searches for a document with a query, he/she expects the most relevant documents to be on top portion of the ranked list. A user usually does not go through the entire search result given by the document retrieval system. So for an IR system it is essential to show the users the most relevant documents in the top portion of the ranked list.

Ranking task is performed by a ranking model/function $f(q,d)$, where $q$ denotes a query and $d$ denotes a document. The traditional ranking models like TF-IDF, BM25 etc. work without a labelled dataset. Supervised machine learning techniques that work on a labelled dataset are being used to create ranking models automatically from labeled training data. In this setting, many features, i.e., traditional ranking models can simultaneously be used to train the ranking model. This is called machine-learned ranking, or rank-learning, or Learning-to-Rank (LTR). LTR is thus a framework that applies supervised machine learning techniques to solve the document ranking problem \cite{ibrahim2015_tf}.

\begin{figure*}[htbp]
\centering
\includegraphics[width=0.7\textwidth]{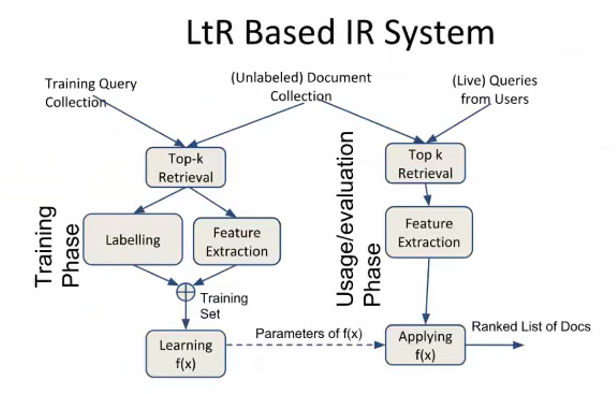}
\caption{Learning-to-Rank based IR system \cite{ibrahim2020empirical}.}
\label{fig:LTR_system}
\end{figure*}

In Figure~\ref{fig:LTR_system} illustrates how an LTR-based IR system works. First, we have to collect a set of queries and a document collection. Using the queries, we then retrieve top $k$ relevant documents using a traditional ranking method such as BM25. We then assign labels (usually an integer in the range 0-4), oftentimes manually, to the documents based on their relevance with respect to a query and extract some features (such as TF-IDF score, BM25 score etc.) from that query-document pair. This way a labelled dataset is created which is ready for being fed into a machine learning algorithm. The algorithm learns a ranking function $f(q,d)$ from this training data. In the testing phase, a query submitted by the user is first used to fetch the top $k$ relevant documents using the same traditional model used prior to the training phase. Then the same features are extracted for this query and the documents. Finally, the learnt ranking model $f(q, d)$ is applied to (re-)rank the retrieved $k$ documents.

LTR framework is being used in a variety of practical ranking tasks \cite{liu2009learning_survey} and is also a vibrant field for theoretical analysis \cite{ibrahim2022understanding}, \cite{shivaswamy2021bias}. Community-based Question-Answering (CQA) task is no exception to that. CQA forums are Internet-based platforms where a user asks questions that other users try to answer. Here the users can also search or browse existing question-answers. If another user has already asked a similar question and received the appropriate answers, a new user with a similar question does not need to ask it again and wait for the users' replies. Instead, the user can search for his/her question in the forum and get the solution instantly. Popular CQA forums include StackOverflow, Yahoo!Answers etc.

The CQA setting is a good application of learning-to-rank paradigm. Here the user searches with a question that acts as a query for the ranking system. He/she expects a relevant question-answer pairs to be in the top positions of the ranked list returned by the system. So here the Question-Answer (QA) pairs are ranked with respect to the query. A real-life example is given below:

\textbf{Query:} ``Unable to parse certain tokens python windows.''

\textbf{Ranked QAs:}

    Q: ``Using libclang parse C Python.'' A: ``Research questions ended exploring library order \ldots''

    Q: ``Parse C array size Python libclang.'' A: ``Currently using Python parse \ldots''

    Q: ``Retrieve function call argument values using libclang.'' A: ``Possible retrieve argument \ldots''

After reviewing the existing literature (as detailed in Section~\ref{sec:related work}) of machine learning-based CQA solutions, we identify some gaps which are as follows: (1) the investigated features cannot be considered extensive or exhaustive, in particular, (a) the use of recurrent neural network-based features and (b) the proper combination of question and/or answer features are not thoroughly investigated, and (2) very few LTR algorithms have been employed so far. Based on these insights, we frame the research question of this study as given below:

\vspace{0.5cm} \textit{What features affect the accuracy of various LTR algorithms in CQA domain and how?}

\vspace{0.5cm} \hspace{-0.5cm} The following sub-questions arise from the above broad question:
\begin{itemize}
    \item \it How can we devise better features for CQA task?
    \item How can we better combine question-question and question-answer features?
    \item \it Which features play the pivotal roles in prediction?
    \item \it Which LTR algorithms perform better in CQA domain?
\end{itemize}


 \hspace{-0.5cm} Below we summarize the contributions made in this paper:

\begin{itemize}
    \item In addition to traditional features like TF-IDF, BM25 etc., we have successfully deployed BERT\footnote{Bidirectional Encoder Representations from Transformers (BERT) is deep learning-based model for natural language processing.}-based feature in CQA domain.
    \item We have investigated not only question-question similarity, but also question-answer similarity features. We have found that features from question-answer pairs are effective for the CQA task when concatenated with question-question features.
    \item We have conducted feature importance analysis and have identified which features have a stronger impact on prediction than others.
    \item We have employed three previously unused yet state-of-the-art LTR algorithms, namely, LambdaLoss, LambdaMART and SERank, in CQA domain. We have evaluated these algorithms on a popular CQA data repository having three different datasets, and compare our performance with existing baselines. In most of the cases, our proposed framework has achieved slightly better accuracy as compared to the existing state-of-the-art algorithms.
\end{itemize}

The rest of the article is organized as follows. Section~\ref{sec:probl formulation} formulates the problem of CQA. Section~\ref{sec:method_overview} details the proposed methodology where we elaborate the framework, investigated features, and the used LTR algorithms. Section~\ref{sec:data and implementation} describes the characteristics of the datasets and implementation details. Section~\ref{sec:results} explains the experimental results and findings. Section~\ref{sec:related work} relates the proposed work with existing literature. Finally, Section~\ref{sec:conclusion} concludes the paper.

\section{Problem Formulation}
\label{sec:probl formulation}
Suppose that we have a set of $M$ questions $q_1, q_2, ..., q_M$ and $k$ number of answers $a_{qi1}, a_{qi2}, ..., a_{qik}$ associated with a question $q_i$. These answers are top rated ones voted by the users of the CQA forum at hand. Now, a new user submits a question $q_u$ to the system. The system is required to show the user a ranked list of similar questions along with their top rated answers. In order to achieve this, the system uses a ranking function $f(.)$ to rank the existing questions according to their relevance with the user's query. From the user's perspective, he/she wants the top-ranked questions to be similar to the submitted one and also the answers to be relevant to his/her information need. So the system, while generating the ranked list, should consider both the (existing) questions and their answers. Therefore, the ranking function $f(.)$ may take the shape $f(q_u, q_i)$, or $f(q_u, a)$, or $f(q_u, \{q_i, a_{qi}\})$ where $a_{qi} = \{a_{qi1}, a_{qi2}, ..., a_{qik}\}$, and $a$ is an answer associated with a question. Sometimes we omit the subscript $i$ to make the notations easier; so in this case $q$ and $a$ means $q_i$ and $a_{qi}$ respectively.

The ground truth label in this setting is usually expressed by the question-question similarity labeled by humans (or by any automated labeling technique), and denoted by an integer number, usually 0 or 1 where 1 means the two questions are similar to each other, and 0 otherwise.

\section{Proposed Methodology}
\label{sec:method_overview}

In order to effectively learn the ranking function $f$, one needs to train this model on a dataset where the feature vectors are computed using both a question and its top answers. However, as detailed in Section~\ref{sec:related work}, many of the existing CQA systems either do not consider answers of a question, or use only the answers without considering the questions. Very few, if at all, studies deal with both the questions and answers to extract features. In this work we investigate as to how we can use features from both a question and its top answers for an LTR algorithm in CQA domain.


\begin{figure*}[h!]
\centering
\includegraphics[width=0.7\textwidth]{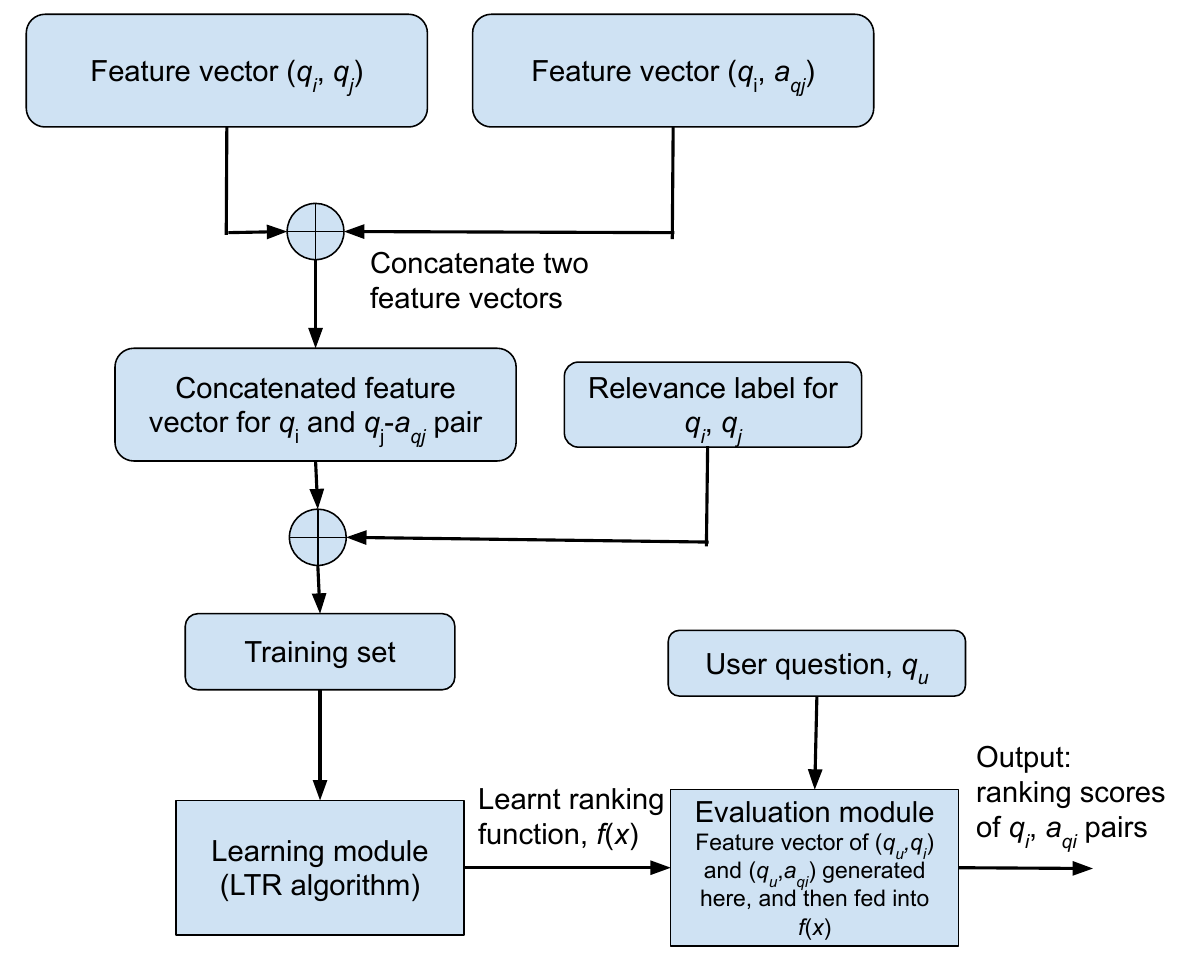}
\caption{Overview of the proposed framework that uses features from both question-question and question-answer pairs.}
\label{fig:main_overview}
\end{figure*}

In Figure~\ref{fig:main_overview} we illustrate the overview of our proposed framework for tackling the CQA problem. As mentioned earlier, a CQA forum dataset usually contains a pool of questions and some top-rated answers associated with each question -- all in natural text format. Treating a question $q_i$ as a user-submitted query, for each $q_i$, we generated two types of traditional IR features\footnote{Recall that examples of IR features include TF-IDF, BM25 etc.}: features extracted from $q_i$ and another question $q_j$ (denoted by $features(q_i, q_j)$), and features extracted from $q_i$ and its top-rated answers (denoted by $features(q_i, a_{qj})$. The reason for taking question-answer features into account is, the questions are oftentimes quite small, and hence only question-question similarity features -- that include mostly the keyword-based similarity -- may not reveal the correct similarity measurement between user's question and answers. We then concatenate these two feature vectors to make it a single one, appended to which is the relevance label of $(q_i, q_j)$, i.e., a label indicating how related these two questions are. Thus for each pair $q_i$ and $q_j$, we generate a feature vector along with a label. These information comprise the training set for the learning phase. In the learning phase, an LTR algorithm is employed that learns a ranking function $f(x)$ from this training set where $x$ is the (concatenated) feature vector. In the evaluation phase, user's question $q_u$ is fed into the system, from which the feature vector ($features(q_u, q_i)$ and $features(q_u, a_{qi})$) is generated. The ranking function thus produces a relevance score for each of the question-answer pairs of the pool, and finally presents a list ranked by these scores.

\subsection{Investigated Features}
\label{subsec:feature description}

As mentioned in Section~\ref{sec:introduction motivation}, a feature value in IR is a relevance score for a document with respect to a query. Accordingly, for query $q$ and document $d$, a feature is a function $f(q,d)$. Below we briefly review the features we use in this investigation:

\paragraph*{Term Frequency:}
Term Frequency (TF) is the number of a word occurred in a document, and is denoted by $TF(t,d)$ where $t$ is a term of the query and $d$ is the document. $TF(t_i,d)$ represents the occurrence of the $i$th term of query $q$ in document $d$.\footnote{Note that in our case we may treat a question as a document.} 

\paragraph*{Inverse Document Frequency:}
Inverse Document Frequency (IDF) is a statistical weight that represents the importance of a term in a collection of documents. If a rare word is present in a document, its IDF score should be higher, and vice versa, because the distinctive power of a rare word with respect to a document is higher. IDF is calculated as $IDF(t) = \log \frac{N}{n(t)}$ 
where $N$ is the total number of documents in the collection, and $n(t)$ is the number of documents containing the term $t$.

\paragraph*{BM25:}
BM25 is another popular model for producing a relevance score for a query and a document. 
It is based on a probabilistic retrieval framework. 
For a given query $q$, with terms, $t_i$, the BM25 score of document $d$ is computed as:
\begin{equation}
\label{eq:bm25}
BM25(d,q) = \sum_{i=1}^{M} \frac{ IDF(t_i) \cdot TF(t_i, d) \cdot (k_1 + 1)} {f(t_i, d) + k_1 \cdot (1 - b + b \cdot \frac{len(d)}{avdl})},
\end{equation}

where $TF(t_i,d)$ and  $IDF(t_i)$ are term frequency and inverse document frequency, $len(d)$ is the number of words of document, $avdl$ is the average document length in the collection, and $k_1$ and $b$ are the tunable parameters.

\paragraph*{Semantic Similarity:}
The semantic similarity of two sentences expresses how similar the two sentences are. We employ a state-of-the-art model called BERT \cite{devlin2018bert} to compute this similarity score. BERT (Bidirectional Encoder Representations from Transformer) language model is based on transformer architecture and is being heavily used for a variety of Natural Language Processing (NLP) tasks since its inception in 2018. Transformer has been well-known for their capability of capturing long-range dependency, and BERT makes a huge leap in that direction. BERT captures a lot of interesting and difficult NLP concepts where previous transformer models did not succeed \cite{rogers2020primer}. 

\begin{figure*}[htbp]
\centering
\includegraphics[width=0.32\textwidth]{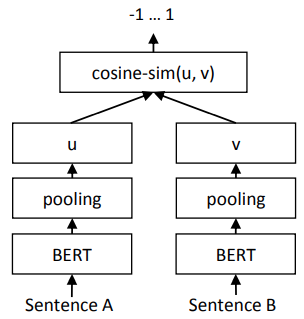}
\caption{Computing cosine similarity of two natural language sentences/documents using BERT model \cite{reimers2019sentence}.}
\label{fig:sentence_embedding}
\end{figure*}

In Figure~\ref{fig:sentence_embedding}, we show how the sentence embedding works and we get a semantic similarity score. First, two sentences are passed into the BERT model. BERT gives us a vectorized representation of an input sentence. After pooling, we get two vectors with identical lengths. We then calculate the cosine similarity between these two vectors which always lies between -1 and 1, where 1 means two sentences are identical, and -1 means they are semantically opposite to each other.

\subsection{Feature Generation}
\label{subsec:feature generation and list}

In this investigation we use a dataset called LinkSO\footnote{The dataset is available at \url{https://sites.google.com/view/linkso}.} where an instance consists of one question and the top two answers. More details of the dataset will be discussed in Section~\ref{sec:data and implementation}.

After performing the preprocessing tasks (like stemming) on the questions and answer texts of the dataset, we extract 35 features in total which are given in Table~\ref{tab:features}. As explained earlier, we extract two types of features: (1) 21 features are extracted from the user's query (denoted by $qid1$) and a corresponding question (denoted by $qid2$) i.e. in the shape $features(qid1, qid2)$ and (2) 14 features are extracted from the user's query and the top-rated answers of $qid2$, i.e., in the shape $features(qid1, answers\_of\_qid2)$. (Note that if multiple answers are in the pool for a single question, we append those answers to make it a single piece of text.) However, some features do not depend on $qid1$ (such as IDF). 



\begin{longtable}[c]{| c | c | c |}
\hline
Feature ID & Feature Description & Stream\\
\hline
\endfirsthead
\hline
Feature ID & Feature Description & Stream\\
\hline
\endhead
1 & Sum of term frequencies (TF)& question \\
\hline
2 & Minimum of term frequencies (TF) & question \\
\hline
3 & Maximum of term frequencies (TF) & question \\
\hline
4 & Average of term frequencies (TF) & question \\
\hline
5 & Variance of term frequencies (TF) & question\\
\hline
6 & Sum of normalized term frequencies (TF) & question\\
\hline
7 & Minimum of normalized term frequencies (TF) & question\\
\hline
8 & Maximum of normalized term frequencies (TF) & question\\
\hline
9 & Average of normalized term frequencies (TF) & question\\
\hline
10 & Variance of normalized term frequencies (TF) & question\\
\hline
11 & Minimum of Inverse document frequencies (IDF) & question\\
\hline
12 & Maximum of Inverse document frequencies (IDF) & question\\
\hline
13 & Average of Inverse document frequencies (IDF) & question\\
\hline
14 & Variance of Inverse document frequencies (IDF) & question\\
\hline
15 & Minimum of TF*IDF & question\\
\hline
16 & Maximum of TF*IDF & question\\
\hline
17 & Average of TF*IDF & question\\
\hline
18 & Variance of TF*IDF & question\\
\hline
19 & Okapi BM25 & question\\
\hline
20 & Cosine similarity score of TF-IDF vectors & question\\
\hline
21 & Semantic similarity using BERT & question\\
\hline
\hline
22 & Sum of normalized term frequencies (TF) & answer\\
\hline
23 & Minimum of normalized term frequencies (TF) & answer\\
\hline
24 & Maximum of normalized term frequencies (TF) & answer\\
\hline
25 & Average of normalized term frequencies (TF) & answer\\
\hline
26 & Variance of normalized term frequencies (TF) & answer\\
\hline
27 & Minimum of Inverse document frequencies (IDF) & answer\\
\hline
28 & Maximum of Inverse document frequencies (IDF) & answer\\
\hline
29 & Average of Inverse document frequencies (IDF) & answer\\
\hline
30 & Variance of Inverse document frequencies (IDF) & answer\\
\hline
31 & Minimum of TF*IDF & answer\\
\hline
32 & Maximum of TF*IDF & answer\\
\hline
33 & Average of TF*IDF & answer\\
\hline
34 & Variance of TF*IDF & answer\\
\hline
35 & Okapi BM25 & answer\\
\hline
\caption{List of investigated features.\label{tab:features}}
\end{longtable}

In IR community it is a common practice to use variations of a single feature like maximum, minimum, variance etc. of TF. This practice is followed in this study.

The features extracted from the question part (i.e.,  $features(qid1, qid2)$) are as follows:
\begin{itemize}
    \item The first 5 features (1-5) are sum, minimum, maximum, average, and variance of term frequencies of the question $qid2$.
    \item Next 5 features (6-10) are sum, minimum, maximum, average, and variance of normalized term frequencies of the question $qid2$.
    \item Next 4 features (11-14) are minimum, maximum, average, and variance of inverse document frequencies of the question $qid2$.
    \item Next 4 features (15-18) are minimum, maximum, average, and variance of TF*IDF. Here normalized term frequency is used as TF.
    \item Feature 19 is the BM25 score of questionc $qid2$.
    \item Feature 20 is the cosine similarity score of tf-idf vector between query $qid1$ and  question $qid2$.
    \item Feature 21 is the cosine similarity score of vectorized representation (produced by BERT) of query $qid1$ and question $qid2$. It is denoted as semantic similarity.
\end{itemize}

The features extracted from the answers (i.e., $features(qid1, answers\_of\_qid2)$) are as follows:
\begin{itemize}
    \item The first 5 features (22-26) are sum, minimum, maximum, average, and variance of normalized term frequencies of the concatenated answers of question $qid2$.
    \item Next 4 features (27-30) are minimum, maximum, average, and variance of inverse document frequencies of the concatenated answers of $qid2$.
    \item Next 4 features (31-34) are minimum, maximum, average, and variance of TF*IDF of concatenated answers of question $qid2$. Here normalized term frequency is used as TF.
    \item Feature 35 is the BM25 score of the concatenated answers of $qid2$.
\end{itemize}

\subsection{Investigated Algorithms}
\label{subsec: algorithm names}

We employ three state-of-the-art LTR algorithms that, to the best of our knowledge, have not been used for the CQA task on LinkSO dataset. Below we present a brief description of the algorithms.

\paragraph*{LambdaLoss:}
LambdaLoss \cite{Wang2018TheLF} is a probabilistic framework  used for metric optimization in ranking tasks. It is based on the popular LambdaRank framework \cite{ibrahim_burges2010ranknet} with some well-defined metric-driven loss functions to reduce the loss over the training data. 

\paragraph*{LambdaMART:}
LambdaMART \cite{wu2008ranking} also uses the LambdaRank framework with a special regression tree called Multiple Additive Regression Trees (MART). MART uses gradient boosted decision trees for the prediction task. LambdaMART uses these trees, but it uses an extra cost function that can be derived from LambdaRank. Using this cost function has shown better results than LambdaRank.

\paragraph*{SERank:}
SERank \cite{wang2020serank} is an effective and efficient method in ranking task. It is a sequence-wise ranking model which uses Squeeze and Excitation network. This network captures cross-document information, which combines scores and sorts a sequence of ranking items. SERank takes an order of documents as input and assigns scores to them together rather than taking a single document at a time. 

\subsection{Detailed Description of the Proposed Framework}
\label{subsec:method detailed}
\begin{figure*}[h!]
\centering
\includegraphics[width=0.7\textwidth]{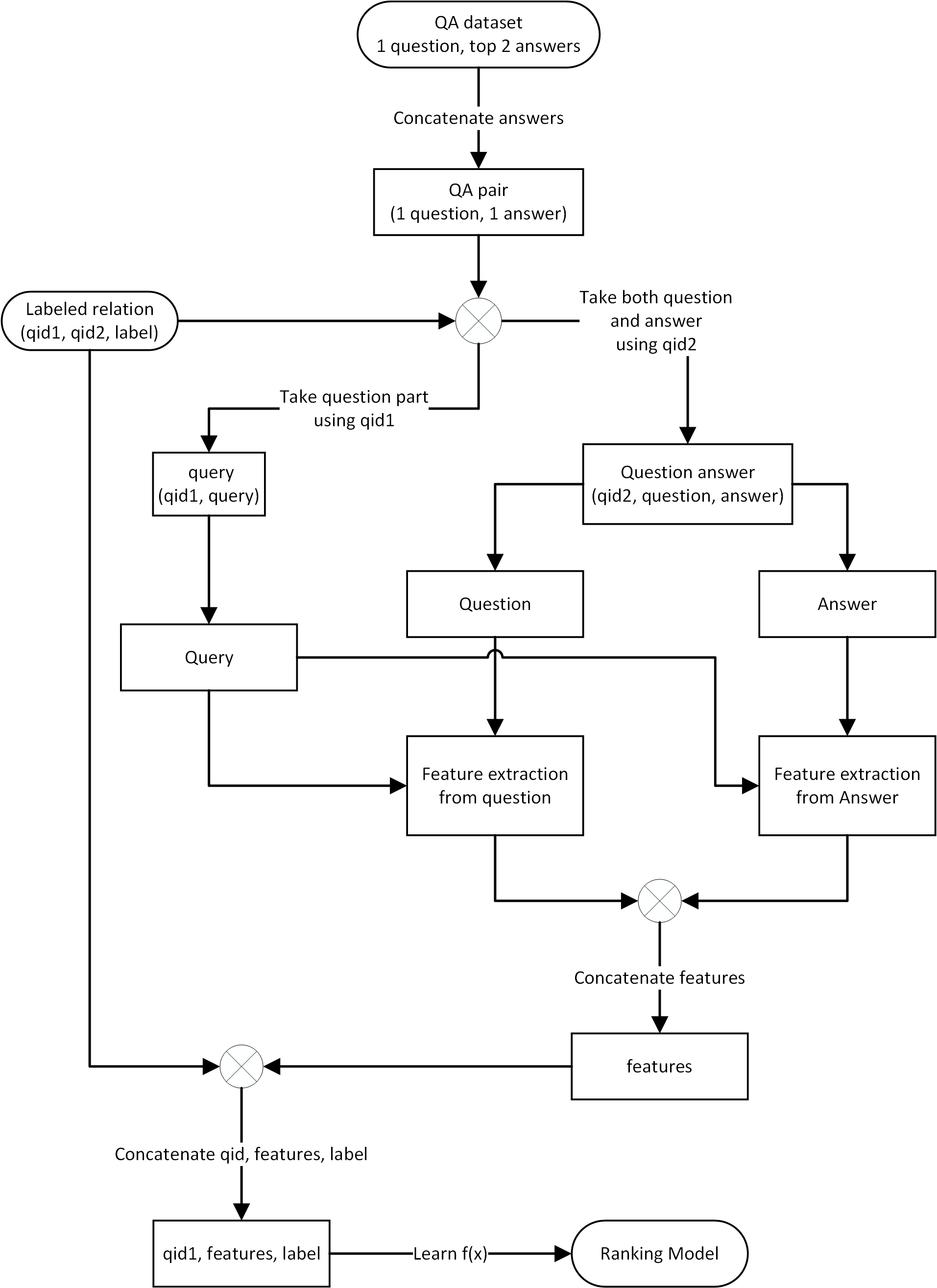}
\caption{Detailed description of proposed framework that uses BERT-based similarity features along with traditional features from both question-question and question-answer pairs.}
\label{fig:main}
\end{figure*}

Now that we have explained all the ingredients of our proposed framework, in Figure~\ref{fig:main} we show a detailed pictorial description of it.

As mentioned earlier, we use a dataset called LinkSO where an instance consists of one question and the top two answers. For each question, we concatenate the two answers to make it one piece of text, thereby having a question-answer pair. The dataset developers provide a label (0 means irrelevant and 1 means relevant) for each question-question pair which indicates how relevant/similar these two questions are. Using this relationship, we then extract the questions using $qid1$, which is going to be used as a query. Each $qid1$ has a labeled relation (relevant or non-relevant) with 30 other questions (denoted by $qid2$). We also extract both the question and answer using $qid2$, which will be used as QA pair for training, validation, and testing phases. Using the query $qid2$, we extract features from the question (21 features) and from the answer (14 features) of these QA pairs independently (cf. Table~\ref{tab:features}). We then concatenate these features to make it a single feature vector (having 35 features in total). We also concatenate the $qid$s and labels with the feature vector for each row which gives us the LibSVM format dataset.\footnote{LibSVM is a popular data format among machine learning practitioners. To know more about LibSVM format, please visit \url{https://www.csie.ntu.edu.tw/~cjlin/libsvmtools/datasets/}.} This builds our processed dataset on which we are going to apply ranking algorithms. We divide these data into three groups for training, validation, and test phases. Using the training data, we learn our ranking model, and then later evaluate the model using test data.

\section{Details of Dataset and Implementation}
\label{sec:data and implementation}
This section discusses the properties of the dataset, implementation details of ranking algorithms, various third-party libraries used and ranking evaluation metrics.


\subsection{Data Description}
StackOverflow\footnote{\url{https://stackoverflow.com/}} is a community-based question answering forum where users ask different questions related to programming and other users try to give proper answers to the questions. We use a labeled QA dataset from StackOverflow dump named LinkSO \cite{liu2018linkso}. LinkSO is a collection of three datasets tagged under the headings ``Java'', ``JavaScript'', and ``Python''. In each category, each question has two top-rated answers. In each category, there are four columns. The first column is the question ID (qid). The next column is the question in text format. The other two columns contain the two top-voted answers, in text format, for that question.

A question may be relevant (denoted by 1) or irrelevant (denoted by 0) to another question. For each question, 30 other questions' similarity (in terms of 1 or 0) are given in the dataset.

In Table \ref{tab:Dataset_statistics} we shown the statistics of LinkSO dataset. Here the number of QA is the corpus size of a particular. 
Train, dev, and test denote the number of training, validation, and testing queries. 




\begin{table*}[htbp]
\begin{center}
\begin{tabular}{ | c |  c | c | c | c | c | c | }
\hline
Dataset Name & \# of QA Pairs & \# of links & \# of queries & train & dev & test\\
\hline
Python & 485,827 & 7,406 & 6,410 & 4,910 & 500 & 1,000\\
\hline
Java & 700,552 & 9,743 & 8,448 & 6,948 & 500 & 1,000\\
\hline
JavaScript & 1,319,328 & 9,444 & 8,069 & 6,569 & 500 & 1,000\\
\hline
Total & 2,505,707 & 26,593 & 22,927 & 18,427 & 1500 & 3,000\\
\hline
\end{tabular}
\vspace{0.3cm}
\caption{\label{tab:Dataset_statistics} Statistics of LinkSO Dataset.}
\end{center}
\end{table*}


\subsection{Implementation Details}

\subsubsection{Hyperparameter Settings of the LTR Algorithms}
For the sake of reproducibility of the experiments, below we describe the parameter settings of the three LTR algorithms:

LambdaLoss: Here we use tensorflow-ranking  library \cite{TensorflowRankingKDD2019}. The parameters we set are as follows: \emph{learning\_rate} = 0.1, \emph{dropout\_rate} = 0.5, \emph{list\_size} = 100, \emph{group\_size} = 1, \emph{num\_train\_steps} = 3000.

LambdaMART: Here we use XGBoost  library \cite{chen2016xgboost} which includes LambdaMART implementation. The parameters we set are: \emph{objective} = (rank:ndcg, map), \emph{eta} = 0.3, \emph{gamma} = 1, \emph{min\_child\_weight} = 0.1, \emph{max\_depth} = 3, \emph{eval\_metric} = (ndcg@5, ndcg@10, map@5, map@10), \emph{tree\_method} = auto, \emph{num\_boost\_round} = 500.

SERank: Here we use SERank implementation from tensorflow-ranking library. The parameters are: \emph{learning\_rate} = 0.1, \emph{dropout\_rate} = 0.5, \emph{list\_size} = 100, \emph{group\_size} = 1, \emph{num\_train\_steps} = 5000, \emph{serank} = True, \emph{query\_label\_weight} = True.

The other parameters have the default values.

\subsubsection{Feature Extraction}
We stem the text data using Python NLTK stemmer. After stemming, we tokenize the question and the two answers. We then concatenate the two answers and thus convert them into a single piece of text.

To extract the features like TF and others, we use the Pandas library. 
For calculating minimum, maximum, mean, and variance, we use the Statistics library package. We use the Counter package for counting document frequencies. For calculating the semantic similarity score, we use the Sentence Transformer library package of Python. For BM25, we set the parameters $k1$ = 1.2, $b$ = 0.75.

We run the experiments on a Windows 11 machine having Intel Core-i7-8700 processor and 32GB RAM. The entire framework is written in Python language.


\subsection{Metrics}
For evaluation metrics, we choose two popular and heavily used IR metrics, namely Normalized Discounted Cumulative Gain (NDCG) and Mean Average Precision (MAP). In particular, MAP@5, MAP@10, NDCG@5, NDCG@10 are used. Since these are well-known metrics in IR community, we do not explain them here; to know details of these metrics, readers may go through \cite{ibrahim2015_tf}.

The TensorFlow-ranking \cite{TensorflowRankingKDD2019} and XGBoost \cite{chen2016xgboost} library provide all the necessary ranking metrics which are used in our experiments.




\section{Experimental Results and Findings}
\label{sec:results}

In this section We analyze the experimental feature set, the importance of any feature, and how this feature set performs on different algorithms.

\subsection{Result Analysis}

\subsubsection{Comparison among Investigated LTR Algorithms}

\begin{table*}[htbp]
\begin{center}
\begin{tabular}{ | c | c | c | c | c | c | c | c | c | c | c | c | c |}
\hline
 \multirow{3}{*}{\backslashbox{Method}{Dataset}} & \multicolumn{4}{c|}{LinkSO-Python} & \multicolumn{4}{c|}{LinkSO-Java}  & \multicolumn{4}{c|}{LinkSO-JavaScript} \\
 \cline{2-13}
 & \multicolumn{2}{c|}{map} & \multicolumn{2}{c|}{ndcg}     & \multicolumn{2}{c|}{map} & \multicolumn{2}{c|}{ndcg}      & \multicolumn{2}{c|}{map} & \multicolumn{2}{c|}{ndcg}\\
 \cline{2-13}
 & @5 & @10 & @5 & @10         & @5 & @10 & @5 & @10  & @5 & @10 & @5 & @10\\
 \hline
 LambdaLoss & 0.438 & 0.464 & 0.495 & 0.555      & 0.450 & 0.476 & 0.512 & 0.570     & 0.489 & 0.514 & 0.550 & 0.600\\
 \hline
 LambdaMART & 0.445 & 0.472 & 0.508 & 0.561    & 0.453 & 0.478 & 0.514 & 0.575    & 0.484 & 0.507 & 0.550 & 0.598\\

 \hline
 SERank & 0.447 & 0.474 & 0.505 & 0.564 & 0.465 & 0.490 & 0.524 & 0.580 & 0.484 & 0.512 & 0.550 & 0.599\\

 \hline
\end{tabular}
\vspace{0.3cm}
\caption{\label{tab:alldata_allmetric} With both question-question and question-answer features: accuracy of all three LTR algorithms on all three datasets of LinkSO.}
\end{center}
\end{table*}

``Python'' is the smallest dataset in LinkSO, and ``JavaScript'' is the largest. Table~\ref{tab:alldata_allmetric} shows the MAP and NDCG values of all three algorithms on all three datasets. 
We see that performance of different algorithms are quite close to each other. On ``Python'' and ``Java'' datasets, LambdaLoss is slightly beaten by the other two, but in ``JavaSrcipt'' dataset, it wins over the others.

\subsubsection{Comparison with Existing Works}

\begin{table*}[htbp]
\begin{center}
\resizebox{\columnwidth}{!}{%
\begin{tabular}{ | c | c | c | c | c | c | c | c |}

\hline
  Work & \multirow{2}{*}{\backslashbox{Method}{Dataset}} & \multicolumn{2}{c|}{LinkSO-Python} & \multicolumn{2}{c|}{LinkSO-Java} & \multicolumn{2}{c|}{LinkSO-JavaScript}\\
 \cline{3-8}
  & & ndcg@5 & ndcg@10 & ndcg@5 & ndcg@10 & ndcg@5 & ndcg@10\\
 \hline
 \multirow{3}{*}{Existing Algorithms \cite{liu2018linkso}} & TF-IDF & 0.301 & 0.360 & 0.282 & 0.352 & 0.315 & 0.378\\
 \cline{2-8}
  & BM25 & 0.320 & 0.384 & 0.321 & 0.382 & 0.344 & 0.412\\
 \cline{2-8}
  & TransLM \cite{xue2008retrieval} & 0.502 & 0.553 & 0.487 & 0.544 & 0.528 & 0.573\\
 \cline{2-8}
  & DSSM \cite{huang2013learning} & 0.461 & 0.519 & 0.443 & 0.500 & 0.461 & 0.519\\
 \cline{2-8}
  & DRMM \cite{guo2016deep} & 0.509 & 0.564 & 0.506 & 0.555 & 0.546 & 0.595\\
 \cline{2-8}
  & aNMM \cite{yang2016anmm} & \textbf{0.514} & \textbf{0.570} & 0.507 & 0.559 & 0.548 & 0.597\\
 \hline
 \multirow{3}{*}{Our Investigation} & LambdaLoss \cite{wang2018lambdaloss} & 0.495 & 0.555 & 0.512 & 0.570 & \textbf{0.550} & 0.600\\
 \cline{2-8}
  & LambdaMart \cite{wu2008ranking} & 0.508 & 0.561 & \textbf{0.514} & \textbf{0.575} & \textbf{0.550} & 0.598\\
 \cline{2-8}
  & SERank \cite{wang2020serank} & 0.505 & 0.561 & 0.512 & 0.570 & \textbf{0.550} & \textbf{0.601}\\

 \hline

\end{tabular}%
}
\vspace{0.3cm}
\caption{\label{tab:comparison} Performance comparison between existing methods and our investigated methods.}
\end{center}
\end{table*}

In Table~\ref{tab:comparison}, we show the performance comparison between existing state-of-the-art algorithms and ours. We see that in general, our methods perform quite good as compared to the state-of-the-art. While on ``Python'' dataset, aNMM method prevails over that of ours, on ``Java'' and ``JavaScript'' datasets, our methods win. In particular, on ``Java'' dataset our methods yield the highest increase in performance.  

In terms of robustness, all three methods of ours demonstrate good performance in all metrics across all datasets. Among the top three performers of the existing baslines (TransLM, DRMM, and aNMM), TransLM is not found to be robust as its performance drastically goes down for NDCG@5 in ``Java'' dataset and NDCG@10 in ``JavaScript'' dataset.

\subsubsection{Importance of BERT-Based Similarity Feature}

\begin{table*}[htbp]
\begin{center}
\begin{tabular}{ | c | c | c | c | c | c | c | c | c | c | c | c | c |}
\hline
 \multirow{3}{*}{\backslashbox{Method}{Dataset}} & \multicolumn{4}{c|}{LinkSO-Python} & \multicolumn{4}{c|}{LinkSO-Java}  & \multicolumn{4}{c|}{LinkSO-JavaScript} \\
 \cline{2-13}
 & \multicolumn{2}{c|}{map} & \multicolumn{2}{c|}{ndcg}     & \multicolumn{2}{c|}{map} & \multicolumn{2}{c|}{ndcg}      & \multicolumn{2}{c|}{map} & \multicolumn{2}{c|}{ndcg}\\
 \cline{2-13}
 & @5 & @10 & @5 & @10         & @5 & @10 & @5 & @10  & @5 & @10 & @5 & @10\\
 \hline
 LambdaLoss & 0.440 & 0.463 & 0.497 & 0.552     & 0.440 & 0.463 & 0.502 & 0.554    & 0.466 & 0.491 & 0.529 & 0.583\\
 \hline
 LambdaMART & 0.439 & 0.465 & 0.498 & 0.557    & 0.435 & 0.461 & 0.502 & 0.560    & 0.469 & 0.493 & 0.531 & 0.583\\
 \hline
 SERank    & 0.451 & 0.456 & 0.507 & 0.563   & 0.446 & 0.472 & 0.507 & 0.564     & 0.473 & 0.498 & 0.534 & 0.588\\
 \hline
\end{tabular}
\vspace{0.3cm}
\caption{\label{tab:withoutBERT_alldata_allmetric} Without BERT-based semantic similarity feature: accuracy of all three LTR algorithms on all three datasets of LinkSO.}
\end{center}
\end{table*}

In order to investigate as to how effective is the BERT-based semantic similarity feature in overall prediction, we conduct the experiments without feature No. 21 of Table~\ref{tab:features}.  Table~\ref{tab:withoutBERT_alldata_allmetric} shows results of all algorithms on all datasets without the BERT-based feature. We can see that without this feature, performance is decreased for almost all the metrics (compare with Table~\ref{tab:alldata_allmetric}), which demonstrates the effectiveness of BERT-based similarity feature.

\subsubsection{Importance of Combining Question-Question and Question-Answer Features}

As mentioned before, most of the existing works use either only question-question features or only question-answer features. In our work we have used both of these two types of features. In order to empirically show that using both indeed helps in prediction, 
Table~\ref{tab:withQQfeatures_alldata_allmetric} shows performance of all three algorithms using only question-question features (i.e., features No. 1-21 of Table~\ref{tab:features}), and 
Table~\ref{tab:withQAfeatures_alldata_allmetric} shows performance using only question-answer features (i.e., features No. 22-35 of Table~\ref{tab:features}). We notice that performance of both of these settings is decreased as compared to the concatenated feature vector we propose (cf. Table~\ref{tab:alldata_allmetric}). This shows that using both question-question and question-answer features in the way we use indeed helps improve performance. Another insight from these experiments is, using only question-question features is better than using only question-answer features. This is, we conjecture, due to the fact that the same answer may be perfect reply to more than one questions, and hence question-question similarity is more important than question-answer similarity.

\begin{table*}[htbp]
\begin{center}
\begin{tabular}{ | c | c | c | c | c | c | c | c | c | c | c | c | c |}
\hline
 \multirow{3}{*}{\backslashbox{Method}{Dataset}} & \multicolumn{4}{c|}{LinkSO-Python} & \multicolumn{4}{c|}{LinkSO-Java}  & \multicolumn{4}{c|}{LinkSO-JavaScript} \\
 \cline{2-13}
 & \multicolumn{2}{c|}{map} & \multicolumn{2}{c|}{ndcg}     & \multicolumn{2}{c|}{map} & \multicolumn{2}{c|}{ndcg}      & \multicolumn{2}{c|}{map} & \multicolumn{2}{c|}{ndcg}\\
 \cline{2-13}
 & @5 & @10 & @5 & @10         & @5 & @10 & @5 & @10  & @5 & @10 & @5 & @10\\
 \hline
 LambdaLoss & 0.413 & 0.437 & 0.468 & 0.522    & 0.424 & 0.446 & 0.478 & 0.531    & 0.455 & 0.478 & 0.513 & 0.567\\
 \hline
 LambdaMART & 0.419 & 0.441 & 0.466 & 0.524    & 0.413 & 0.436 & 0.465 & 0.521    & 0.448 & 0.470 & 0.507 & 0.558\\
 \hline
 SERank & 0.423 & 0.443 & 0.478 & 0.528      & 0.429 & 0.450 & 0.485 & 0.535     & 0.455 & 0.479 & 0.514 & 0.567\\
 \hline
\end{tabular}
\vspace{0.3cm}
\caption{\label{tab:withQQfeatures_alldata_allmetric} With question-question features only: accuracy of all three LTR algorithms on all three datasets.}
\end{center}
\end{table*}

\begin{table*}[htbp]
\begin{center}
\begin{tabular}{ | c | c | c | c | c | c | c | c | c | c | c | c | c |}
\hline
 \multirow{3}{*}{\backslashbox{Method}{Dataset}} & \multicolumn{4}{c|}{LinkSO-Python} & \multicolumn{4}{c|}{LinkSO-Java}  & \multicolumn{4}{c|}{LinkSO-JavaScript} \\
 \cline{2-13}
 & \multicolumn{2}{c|}{map} & \multicolumn{2}{c|}{ndcg}     & \multicolumn{2}{c|}{map} & \multicolumn{2}{c|}{ndcg}      & \multicolumn{2}{c|}{map} & \multicolumn{2}{c|}{ndcg}\\
 \cline{2-13}
 & @5 & @10 & @5 & @10         & @5 & @10 & @5 & @10  & @5 & @10 & @5 & @10\\
 \hline
 LambdaLoss & 0.328 & 0.357 & 0.383 & 0.450     & 0.304 & 0.333 & 0.363 & 0.430    & 0.324 & 0.353 & 0.381 & 0.447\\
 \hline
 LambdaMART & 0.329 & 0.358 & 0.388 & 0.448    & 0.312 & 0.341 & 0.369 & 0.432    & 0.323 & 0.352 & 0.380 & 0.442\\
 \hline
 SERank  & 0.335 & 0.363 & 0.390 & 0.456    & 0.330 & 0.355 & 0.389 & 0.447   & 0.341 & 0.368 & 0.399 & 0.459\\
 \hline
\end{tabular}
\vspace{0.3cm}
\caption{\label{tab:withQAfeatures_alldata_allmetric} With question-answer features only: accuracy of all three LTR algorithms on all three datasets.}
\end{center}
\end{table*}

\subsection{Feature Importance Analysis}

To find out which features play pivotal roles in predicting the relevance labels, we analyze the importance of the investigated features using the gain-based technique of tree-ensemble models. 

\paragraph*{Gain:}
This feature analysis method works in tandem with tree-based algorithms such as XGBoost \cite{chen2016xgboost}. In a tree, a split at a node is performed based on a particular feature. If we split a leaf into two leaves, the feature gains the following score:
\begin{equation}
\label{eq:gain}
Gain = \frac{1}{2} \left[\frac{G_L^2}{H_L+\lambda}+\frac{G_R^2}{H_R+\lambda}-\frac{(G_L+G_R)^2}{H_L+H_R+\lambda}\right] - \gamma
\end{equation}
The first and second parts of the formula are the scores of the new left and right leaves respectively. The third part is the score on the original (i.e., parent) leaf. The fourth part is regularization on the additional leaf. Also, $\gamma$ and $\lambda$ are regularization parameters. $G_L, G_R$ are gains of left and right leaves, and $H_L, H_R$ are the heights of these leaves respectively.  The higher value of gain of a feature, the more important the feature is for predicting the labels.

We now measure the importance of each of our investigated features using the above-mentioned method. We choose to use XGBoost for analyzing the feature importance since it has been found to be effective in our experiments across the three datasets. We conduct both regression tree and classification tree based analyses.


\begin{figure*}[htbp]
\centering
\subfigure[Python]{\label{fig:python_reg}\includegraphics[width=0.32\textwidth]{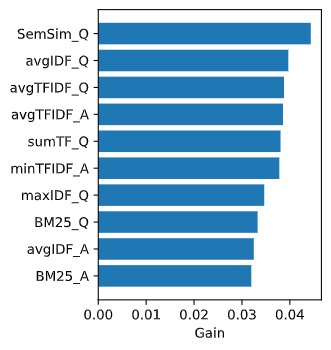}}
\subfigure[Java]{\label{fig:java_reg}\includegraphics[width=0.32\textwidth]{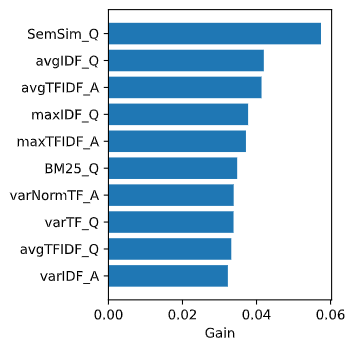}}
\subfigure[JavaScript]{\label{fig:javascript_reg}\includegraphics[width=0.32\textwidth]{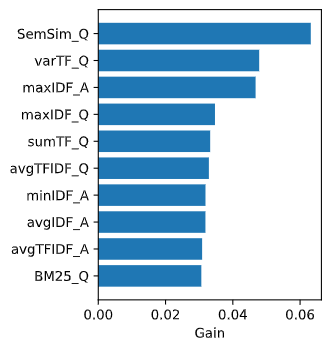}}

\subfigure[Python]{\label{fig:python_class}\includegraphics[width=0.32\textwidth]{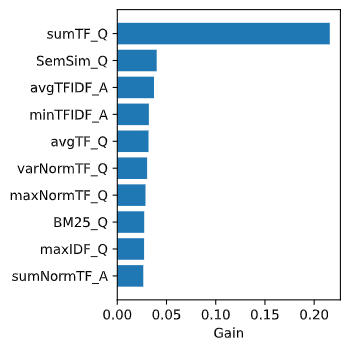}}
\subfigure[Java]{\label{fig:java_class}\includegraphics[width=0.32\textwidth]{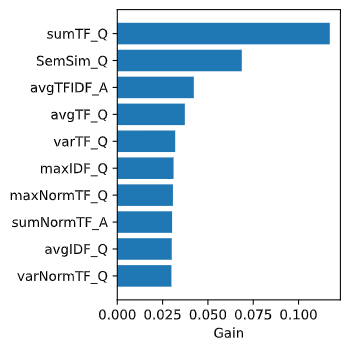}}
\subfigure[JavaScript]{\label{fig:javascript_class}\includegraphics[width=0.32\textwidth]{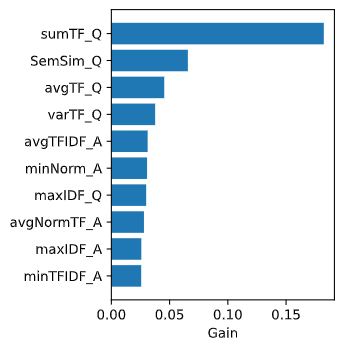}}

\caption{Regression-based (top row) and classification-based (bottom row) feature importance scores of top 10 features for each dataset. `Q' and `A' at the end of feature names indicate question and answer features respectively (cf. Table~\ref{tab:features}).}
\label{fig:reg_gain}
\end{figure*}

In the top row of Figure~\ref{fig:reg_gain} we show the top 10 features for each dataset with regression tree-based analysis. Here, we can see that queries' semantic similarity with the question has the most impact on ranking tasks in each dataset. We can also observe that features extracted from answers also have significant impact.




In the bottom row of Figure~\ref{fig:reg_gain} we show the top 10 features for each dataset with classification tree-based analysis. Here, we can see that sum of term frequencies of the question and the semantic similarity feature highly affect the ranking performance in each dataset. The average TF-IDF of the answer also has a significant impact.


\subsection{Overall Findings}

Below are the three major findings of this study: 

\begin{itemize}
    \item In addition to the traditional features such as TF-IDF, BM25 etc. it is important to capture semantic similarity between the question-question pairs. In our investigation, the BERT-based semantic similarity feature is found the be effective -- both in experiments and in feature importance analysis.
    \item Existing CQA tasks should focus on both question-question and question-answer features since our experiments reveal that features extracted from answers have a significant impact on performance.
    \item Most of the existing baseline algorithms used neural network-based rank-learning algorithms (e.g. DSMM \cite{huang2013learning}, DRMM \cite{guo2016deep}, aNMM \cite{yang2016anmm}). Our findings suggest that tree-ensemble rank-learning algorithms (e.g. LambdaMART) would demonstrate similar level of performance and robustness.
\end{itemize}



\section{Related Works}
\label{sec:related work}



Yang et al. \cite{yang2016anmm} employ attention-based neural network to model $f(q_u, a)$, i.e., to learn a scoring function between user query and answers from the collection. The authors test their algorithm (named aNMM) on TREC QA data with two sets of experiments. Firstly, they do not use any manual features like TF, BM25 etc., rather use only the deep learning model based on word embeddings of question and answer. Secondly, they use the scores of aNMM as a feature along with some other traditional features like TF-IDF and BM25 and then use a rank-learning algorithm.


Martino et al. \cite{da2016learning} model $f(q_u, q)$ using some advanced features but no deep learning methods. The authors evaluate their model on SemEval 2016 Task 3 dataset\footnote{\url{https://alt.qcri.org/semeval2016/task3/}}. \cite{da2016learning} work on similar question retrieval domain where the authors propose an approach which depends on three types of features to retrieve the similarity between the previous and the asked question in the forum. The three features are: text similarities, PTK similarity, and Google rank. 
Ngueyn et al. \cite{nguyen2016learning} also use the same dataset to model $f(q_u, q)$ with RankSVM algorithm.

In CQA forums, there can be missing links between two similar questions. This may hamper the relevance judgment for evaluating the ranking task. Liu et al. \cite{liu2018linkso} propose a dataset called LinkSO which resolves this issue. The authors conduct an empirical investigation on the performance of some existing state-of-the-art algorithms -- both machine learning-based and traditional ones -- on their proposed dataset. They find their dataset is well-suited for learning-based systems.

Question-Answer Topic Model (QATM) is proposed by Ji et al. \cite{ji2012question} to solve the issue of lexical gap between user's query and questions in the dataset. It is used to learn the hidden topics in question-answer pairs to ease the lexical gap problem. The authors assume that a question and its paired answers share the same concept distribution. They work with Yahoo!Answers forum dataset and report that combining latent semantic topics of both question and answer parts significantly improves similar question retrieval and ranking.


Several works use deep learning technique along with traditional features like BM25, IDF etc. \cite{ibrahim_severyn2015learning},
\cite{ibrahim_wang2015long}, \cite{ibrahim_yu2014deep}.

\subsection{LSTM-Based Models}
Tay et al. \cite{tay2017learning} models $f(q_u, a)$ using LSTM architecture of deep learning to capture long term dependencies, coupled with holographic dual method. The authors use no other features except the ones produced by the deep learning model. TREC QA and Yahoo QA datasets are used for evaluation. Their model outperforms many variants of deep learning architectures. One significant difference between document retrieval and similar question retrieval is that questions and answers are often much shorter than usual documents which makes the designing of the features challenging. 

Pei et al. \cite{pei2021attention} use bidirectional LSTM model on word embeddings of question and answer. The authors use no additional features and evaluate their algorithm on LinkSO and AskUbuntu QA datasets. 
Their proposed Attention-based Sentence pair Interaction Model (ASIM) utilizes the attention mechanism to grab the semantic interaction information. 

Nassif et al. \cite{nassif2016learning} model $f(q_u, a)$ on the same dataset with stacked bidirectional LSTM and MLP. The authors build a model that outputs the representation of the vector form of question-question or question-answer pairs and computes their semantic similarity scores. Using this score, the system ranks the questions in a question's list and answers in an answer's list according to their semantic relatedness.

Othman et al. \cite{othman2020improving} address some significant challenges of CQA domain. The challenges are shortness of the questions and word mismatch problems. Users often ask the same question using a different set of words. The authors propose a method that combines a Siamese network and LSTM networks augmented with an attention mechanism.

\subsection{Transformer and BERT-Based models}

Tu et al. \cite{ibrahim_van2021deep} model $f(q_u, q)$ using BERT-based learning module along with some other word-matching features (such as direct word-embeddings, word phrases based on skip-gram etc.). The authors do not use any traditional IR features such as BM25 or TF-IDF. The authors use the SemEval 2016 dataset for evaluation.

Sen et al. \cite{ibrahim_sen2020support} also model $f(q_u, a)$ using BERT deep learning model but in a transfer learning setting. The authors evaluate their model on a QA dataset (known as MSDN dataset) taken from Microsoft Developer Network\footnote{\url{https://docs.microsoft.com/en-us/}}.

Shao et al. \cite{ibrahim_shao2019transformer} employ both transformer and bidirectional LSTM models to learn $f(q_u, a)$ using WikiQA dataset\footnote{\url{https://deepai.org/dataset/wikiqa}}. Almiman et al. \cite{ibrahim_almiman2020deep} model $f(q_u, a)$ using BERT-based and lexical features for Arabic question-answering task. Zahedi et al. \cite{ibrahim_zahedi2020hca} employ an end-to-end Hierarchical Compare Agregate model for CQA task. The authors model $f(q_u, q)$ using  in a transfer learning setting.

We see from the above discussion that the existing works either work on question-question similarity-based features, or only question-answer similarity-based features. But to the best of our knowledge, no work addressed transformer-based similarity features along with other traditional features (like TF-IDF, BM25), and moreover, no work used $f(q_u, q)$ and $f(q_u, a)$ in a linear, concatenated manner. The main novelty of our work lies here: we extract features from both questions and answers but we do it separately, and then concatenate the features. Also, we introduce in the CQA domain the BERT-based cosine similarity features.

\section{Conclusion}
\label{sec:conclusion}
In this investigation we have looked into the the broad area of feature engineering for community question answering domain. We have engineered a BERT-based feature to capture semantic similarity between question-question pair. We have combined both question-question and question-answer features in a novel way. While testing with previous unused state-of-the-art rank-learning algorithms, both of our proposed concepts are found to be effective. The experimental results also show that performance of our approaches are competitive and robust as compared to existing baseline performance. 


\section*{Funding Information}
The authors hereby declare that there was no financial support for this work that could have influenced the outcome of this research.

\section*{Conflict of Interest}
The authors hereby declare that there is no known conflict of interest associated with this article.

\section*{Author Contribution}
\hspace{0.45 cm} Nafis Sajid: Planning, investigation, coding, writing original draft.

Md. Rashidul Hasan: Planning, investigation, coding, writing original draft.

Muhammad Ibrahim: Planning, investigation, writing and reviewing original draft, supervision.

 \bibliographystyle{splncs04}
%
\bibliography{ref}
\end{document}